\def\year{2022}\relax
\documentclass[letterpaper]{article} 
\usepackage{aaai22}  
\usepackage{times}  
\usepackage{helvet}  
\usepackage{courier}  
\usepackage[hyphens]{url}  
\usepackage{graphicx} 
\def\UrlFont{\rm}  
\usepackage{natbib}  
\usepackage{caption} 
\DeclareCaptionStyle{ruled}{labelfont=normalfont,labelsep=colon,strut=off} 
\usepackage{algorithm}

\usepackage{newfloat}
\usepackage{listings}
\floatstyle{ruled}
\newfloat{listing}{tb}{lst}{}
\floatname{listing}{Listing}
\title{Improving Controllability of Educational Question Generation by Keyword Provision}
\author {
    Ying-Hong Chan,
    Ho-Lam Chung,
    Yao-Chung Fan
}
\affiliations {
     Department of Computer Science and Engineering \\
    National Chung Hsing University, \\ 
    Taichung, Taiwan \\
    harry831120@gmail.com, voidful.stack@gmail.com, yfan@nchu.edu.tw
}

\usepackage{bibentry}

\usepackage{balance}
\usepackage{booktabs}
\usepackage{graphicx}
\usepackage{amsmath}
\usepackage{url}
\usepackage{amssymb}
\usepackage{multirow}
\usepackage{algpseudocode}
\usepackage{xcolor}
\usepackage{enumitem}

\begin{document}

\maketitle

\begin{abstract}
Question Generation (QG) receives increasing research attention in NLP community. One motivation for QG is that QG significantly facilitates the preparation of educational reading practice and assessments. While the significant advancement of QG techniques was reported, current QG results are not ideal for educational reading practice assessment in terms of \textit{controllability} and \textit{question difficulty}. This paper reports our results toward the two issues. First, we report a state-of-the-art exam-like QG model by advancing the current best model from 11.96 to 20.19 (in terms of BLEU 4 score). Second, we propose to investigate a variant of QG setting by allowing users to provide keywords for guiding QG direction. We also present a simple but effective model toward the QG controllability task.  Experiments are also performed and the results demonstrate the feasibility and potentials of improving QG diversity and controllability by the proposed keyword provision QG model.
 
\end{abstract}

\section{Introduction}
Question generation (QG), taking a passage and an answer phrase as input and generating a context-related question as output, has received tremendous interests in recent years \cite{zhou2017neural, zhao2018paragraph, du2017learning, chan2019recurrent,dong2019unified,bao2020unilmv2}. 

One motivation of developing QG technology is to facilitate teachers in preparation of reading comprehension assessments. While significant QG quality were reported, we find the following limitations for integrating the current QG models into practical educational usage scenario. 

First, the current QG model suffers from the model controllability concern. This concern is better seen by an example. In Table \ref{tab:CQGexample}, we show two questions ($Q_1$ and $Q_2$). Both of them are eligible with respect to the passage and answer. The issue is that we have no way to control the QG model to generate which question. Therefore, the question generated might not be an user expected result, which lowers user experiences in practical educational preparation scenario.

\begin{table}[t]
\small
\resizebox{\linewidth}{!}{
\begin{tabular}{|p{0.12\columnwidth}|p{0.88\columnwidth}|}
\hline 
Context & At the age of 12, \textbf{Christopher Hirata} already worked on college-level courses, around the time most of us were just in the 7th grade. At the age of 13, this gifted kid became the youngest American to have ever won the gold medal in the International Physics Olympiad. At the age of 16, he was already working with NASA on its project to conquer planet mars. After he was awarded the Ph.D. at Princeton University, he went back to California institute of technology. The next person with very high IQ is Albert Einstein. With an IQ between 160 and 190, Albert Einstein is the genius behind the theory of relativity, which has had great impact on the world of science. He possessed such an amazing ability that after his death, researchers were eager to preserve and make research on his brain in search for clues to his exceptional brilliance, which to this day, has remained a mystery. \\ \hline\hline
Answer & Christopher Hirata \\ \hline\hline
$Q_1$ & Who once worked on the project to conquer planet mars? \\ \hline
$Q_2$ & Who was the youngest American to have ever won the gold medal in the International Physics Olympiad?
 \\ \hline
\end{tabular}
}

\caption{An Example for QG Model Controllability Concern: Note that $Q_1$ and $Q_2$ are both eligible with respect to the context passage and answer. We would like to note that we have no way to control which question to generate.}
\label{tab:CQGexample}\vspace{-5mm}
\end{table}

Second, questions generated by existing QG models are too simple (in terms of assessment difficulty) for advance educational reading practice assessment. Current data-driven QG models are trained with factoid QA datasets (e.g., SQuAD \cite{rajpurkar2016squad} or NewsQA \cite{trischler2016newsqa}), and therefore generate factoid questions, which are too simple for advanced reading practice assessment. 

In this paper, we report our results toward the two problems. First, for the educational QG quality issue, we investigate training QG models with exam-like datasets (e.g., RACE \cite{lai2017race}). Moreover, we investigate the employment of pre-trained language models for exam-like QG. Our experiment results show that the language model employment significantly advances the state-of-the-art result reported by \cite{jia2020eqg} from 11.96 to 20.19 (in terms of BLEU 4 score).

For the controllability issue, we propose to address a new QG setting variant (called CQG, Controllable Question Generation), which allows users to guide the QG direction by indicating keywords expected to be included in the generation result. Specifically, the input to CQG is a triple (\textit{``passage''}, \textit{``answer''}, \textit{``a set of keywords''}) rather than a ``\textit{passage}'' and ``\textit{answer}'' pair (a common QG setting was). In this paper, we approach CQG by introducing KPQG (Keyword Provision Question Generation) model.

\section{Educational QG based on Pre-trained LM}
In this section, we report our results on the employment of pre-trained language models (PLM) for educational QG. We experiment with two existing text generation alternatives (Masked-LM QG Generation and Seq2Seq Generation). 

\subsection{Masked-LM Generation}
We train a Masked-LM generation model $\mathbb{M}()$ taking a context paragraph $C$, answer $A$, and the previous generated tokens $q_1, ..., q_{i-1}$ and as input and outputting a target $q_i$ in an auto-regressive manner.
\begin{align*}
& \mathbb{M}(C\texttt{[S]}A\texttt{[M]}) \rightarrow q_1,\\
& \mathbb{M}(C\texttt{[S]}A\texttt{[S]}q_1\texttt{[S]}\texttt{[M]}) \rightarrow q_2,\\
& \mathbb{M}(C\texttt{[S]}A\texttt{[S]}q_1\texttt{[S]}q_2\texttt{[S]}\texttt{[M]}) \rightarrow q_3,\\
& ... 
\end{align*}

\subsection{Seq2Seq Generation} 
We train a seq2seq model $\mathbb{M}()$ taking a context paragraph $C$ and an answer $A$ as input and predicting a sequence of question tokens $\{q_1,q_2...q_{|Q|}\}$ as output. Specifically, we have
\[\mathbb{M}(C\texttt{[S]}A)\rightarrow q_1, q_2,...q_{|Q|}\]

\begin{figure*}[t]
    \centering
    \includegraphics[width=\textwidth]{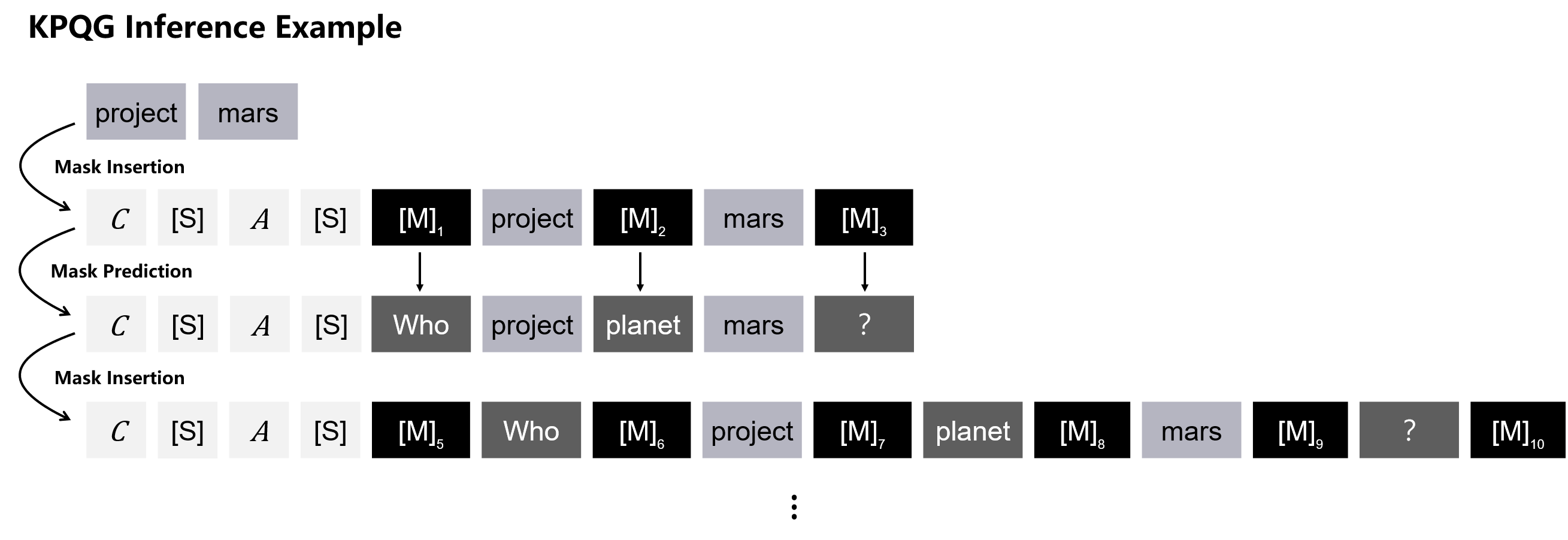}
    \caption{KPQG Mask Insertion and Prediction}
    \label{fig:E_KPQG.png}
\end{figure*}

\subsection{Experiment Results}
\label{sec:sqg-analyze}

\begin{table}[t]
\centering
\begin{tabular}{|l|l|l|l|}
\hline
 & Train & Test & Dev \\
\hline
 \# of instances & 17445 & 950 & 1035 \\
\hline
\end{tabular}
\caption{\textbf{EQG-RACE} Dataset statistics}
\label{table:data num}\vspace{-4mm}
\end{table}

\begin{table*}[t!]
\centering
\begin{tabular}{|c|c|c|c|c|c|c|}
\hline
{\bf Model} & {\bf BLEU 1} & {\bf BLEU 2} & {\bf BLEU 3} & {\bf BLEU 4} &{\bf ROUGE-L} & {\bf  METEOR} \\ 
\hline
{\cite{jia2020eqg}}  & 35.10 & 21.08 & 15.19 & 11.96 & 34.24 & 14.94 \\  
{BERT-QG} & 43.37 & 29.53 & 22.25 & 17.54 & 44.26 & 20.47 \\   
{RoBERTa-QG} & 46.37 & 32.15 & 24.34 & 19.21 & 46.96 & 22.32\\   
{BART-QG} & 46.78 & 32.30 & 24.53 & 19.39 & 47.00 & 22.22\\   
{DeBERTa-QG} & \bf{47.16} & \bf{32.81} & \bf{25.18} & \bf{20.19} & \bf{47.33} & \bf{22.55} \\  
 
\hline
\end{tabular}
\caption{Performance Comparison}
\label{table:race_s_kpqg_score}
\end{table*}

We present the evaluation results for our models on EQG-RACE \cite{jia2020eqg} dataset. Table \ref{table:data num} summarizes statistics for the datasets. 

We implement the Masked-LM QG architecture with BERT \cite{devlin2018bert}, RoBERTa \cite{liu2019roberta}, and DeBERTa \cite{he2020deberta}, and the Seq2Seq QG architecture with BART \cite{lewis2019bart}. All the pre-trained models were provided by \cite{wolf2020huggingfaces} for our model training. The  evaluation is based on the script  released  by  (Du,  Shao,  and Cardie 2017). The evaluation script includes BLEU 4 (Pap-ineni et al. 2002), METEOR (Denkowski and Lavie 2014) and  ROUGE  (Lin  2004)  evaluation  metrics. 


Table \ref{table:race_s_kpqg_score} shows the comparison results. We also list the state-of-the-art results reported by \cite{jia2020eqg}. We see that the PLM employment significantly improves the performance of educational QG. Among them, DeBERTa-QG advances the SOTA result from 11.96 to 20.19 (in terms of BLEU 4 score).

Furthermore, we have the following observations from the experiment results. First, all models are able to generate fluent questions as reported in the QG literature. Second, the inferior performance (in terms of token scores) comes from insufficient answer information. Directly taking the answers as input for QG lefts too much freedom and suffers from insufficient answer information for guiding QG direction, and therefore the observed inferior performance in this token score comparison. This observation also responses to the QG model controllability concern discussed in the introduction section.

Generating questions not expected by users might lower the user experiences in practical educational preparation scenario. Therefore, in the next section, we introduce KPQG, which guides the direction of question generation through the given keywords, so as to address the problem of controllability.




\section{Keyword Provision Question Generation}
In this section, we present Keyword Provision Question Generation (KPQG) scheme for guiding QG generation.

Note that we assume that a set of keywords are given by users in addition to the context paragraph and answer (as in the common QG setting).  

\paragraph{KPQG Inference}
For a given keyword token sequence [$k_1$, $k_2$, $k_3$], we align the input sequence $X_0$ to the PLM models by inserting \texttt{[M]} tokens. Specifically, 


\[ 
X_{0} = [C \mathrm{\texttt{[S]}} A  \mathrm{\texttt{[S]}} \mathrm{\texttt{[M$_1$]}} k_{1}  \mathrm{\texttt{[M$_2$]}} k_{2}
\mathrm{\texttt{[M$_3$]}}]  \]

We leverage Masked-LM generation to predict suitable tokens for \texttt{[M]} tokens. After the prediction, we recursively insert and predict the \texttt{[M]} tokens in the same manner. At each iteration, we align the input sequence by inserting \texttt{[M]} before and after all given/generated tokens. The iteration continues till all masked tokens becomes \texttt{[S]}. 

As a concrete example, please refer to the example shown in Figure \ref{fig:E_KPQG.png}. Two keywords (\texttt{project} and \texttt{mars}) are given in this example. At Iteration 0, we have three inserted \texttt{[M]} tokens, and the predicted results are ``\texttt{Who}'', ``\texttt{planet}'', and ``\texttt{?}''. And, at Iteration 1, we align the input sequence $X_1$ by inserting \texttt{[M]} before and after all given/generated tokens, as shown in Figure \ref{fig:E_KPQG.png}.





\paragraph{KPQG Training}
The KPQG is trained to predict a masked token before/after the input/generated keyword tokens. Under this goal, the challenge is that which tokens in a given question sentence should be masked for model training. Randomly masking for model training is not a solution, as tokens in a question sentence has different importance. 

We explore the idea of learning to predict important word first by employing a QA model to assess the importance of tokens in a given $Q$. Our idea is that if masking some token $q_i$ from a question sentence [$q_1, ..., q_{|Q|}$] leads to to decreased QA model performance, then $q_i$ is considered to be an important one with respect to the QA pairs. Therefore, for a given $Q$, we iteratively replace all tokens in $Q$ with a \texttt{[PAD]} special token in a one-at-a-time manner. 

For example, for a question ``how is the weather today?'', we have the following \textit{padded} question sentences.

\begin{itemize}
    \item \texttt{[PAD]} is the weather today?
    \item how \texttt{[PAD]} the weather today?
    \item how is \texttt{[PAD]} weather today?
    \item how is the \texttt{[PAD]} today?
    \item how is the weather \texttt{[PAD]} ?
    \item how is the weather today \texttt{[PAD]}
\end{itemize}

We then pose the sentences to a QA model for answer prediction, and we estimate the importance of a keyword through the model’s confidence in answer prediction. The lower the confidence, the higher the importance of the masked token. 

Then, we create training data for KPQG based on the token importance. In Table \ref{tab:E-KPQG}, we show an example. Assume that the importance of a question sentence [$q_1$, ..., $q_9$] is $[q_4, q_6, q_2,  q_5, q_3, q_1, q_9, q_7, q_8]$ (from high to low).

We create six training instances as shown in Table \ref{tab:E-KPQG}. The first training instance aims instruct KPQG model to predict the most important word (i.e., $q_4$) based on only Context Paragraph $C$ and Answer $A$. Thus, the label of the \texttt{[M]} token is set to $q_4$. 

\[
\mathbb{M}(
C \mathrm{\texttt{[S]}} A  \mathrm{\texttt{[S]}}  \mathrm{\texttt{[M]}}) \rightarrow q_4
\]

Similarly, the second training instance is set to predict $q_2$ and $q_6$ as follows. 

\[
\mathbb{M}(
C \mathrm{\texttt{[S]}} A  \mathrm{\texttt{[S]}}  \mathrm{\texttt{[M]}} q_{4} \mathrm{\texttt{[M]}}) \rightarrow q_2, q_6
\]

Likewise, we have 
\[
\mathbb{M}(
C \mathrm{\texttt{[S]}} A  \mathrm{\texttt{[S]}}  \mathrm{\texttt{[M]}} q_{2} \mathrm{\texttt{[M]}} q_{4} \mathrm{\texttt{[M]}} q_{6} \mathrm{\texttt{[M]}}) \rightarrow q_1, q_3, q_5, q_9
\]

Please refer to the complete training instances in Figure \ref{tab:E-KPQG}.

\begin{table*}[t]
\small
\centering
\resizebox{\textwidth}{!}{
\begin{tabular}{|l|l|l|}
\hline
 & $X_i$ & Labels for \texttt{[M]} \\ \hline
i=0 &  $C$ $\mathrm{\texttt{[S]}}$ $A$ $\mathrm{\texttt{[S]}}$ $\mathrm{\texttt{[M]}}$ & $q_{4}$ \\ \hline

i=1 &  $C$ $\mathrm{\texttt{[S]}}$ $A$ $\mathrm{\texttt{[S]}}$ $\mathrm{\texttt{[M]}}$ $q_{4}$ $\mathrm{\texttt{[M]}}$ & $q_{2}$ $q_{6}$ \\ \hline

i=2 &  $C$ $\mathrm{\texttt{[S]}}$ $A$ $\mathrm{\texttt{[S]}}$ $\mathrm{\texttt{[M]}}$ $q_{2}$, $\mathrm{\texttt{[M]}}$ $q_{4}$ $\mathrm{\texttt{[M]}}$ $q_{6}$ $\mathrm{\texttt{[M]}}$ 
& $q_{1}$ $q_{3}$ $q_{5}$ $q_{9}$  \\ \hline

i=3 &   $C$ $\mathrm{\texttt{[S]}}$ $A$ $\mathrm{\texttt{[S]}}$ $\mathrm{\texttt{[M]}}$ $q_{1}$ $\mathrm{\texttt{[M]}}$ $q_{2}$ $\mathrm{\texttt{[M]}}$ $q_{3}$ $\mathrm{\texttt{[M]}}$ $q_{4}$ $\mathrm{\texttt{[M]}}$ $q_{5}$ $\mathrm{\texttt{[M]}}$
$q_{6}$ $\mathrm{\texttt{[M]}}$ $q_{9}$ $\mathrm{\texttt{[M]}}$  &
 $\mathrm{\texttt{[S]}}$ $\mathrm{\texttt{[S]}}$  $\mathrm{\texttt{[S]}}$ $\mathrm{\texttt{[S]}}$
$\mathrm{\texttt{[S]}}$ $\mathrm{\texttt{[S]}}$ $q_{7}$ $\mathrm{\texttt{[S]}}$ \\ \hline

i=4 &  $C$ $\mathrm{\texttt{[S]}}$ $A$ $\mathrm{\texttt{[S]}}$ $q_{1}$ $q_{2}$ $q_{3}$ $q_{4}$
$q_{5}$ $q_{6}$ $\mathrm{\texttt{[M]}}$ $q_{7}$ $\mathrm{\texttt{[M]}}$ $q_{9}$
& $\mathrm{\texttt{[S]}}$ $q_{8}$ \\ \hline

i=5 &  $C$ $\mathrm{\texttt{[S]}}$ $A$ $\mathrm{\texttt{[S]}}$ $q_{1}$ $q_{2}$ $q_{3}$ $q_{4}$ $q_{5}$ $q_{6}$ $q_{7}$ $\mathrm{\texttt{[M]}}$ $q_{8}$ $\mathrm{\texttt{[M]}}$ $q_{9}$
&  $\mathrm{\texttt{[S]}}$ $\mathrm{\texttt{[S]}}$ \\ \hline

\end{tabular}
}
\caption{ The training instance creation example: the importance of a question sentence [$q_1$, ..., $q_9$] is $[q_4, q_6, q_2,  q_5, q_3, q_1, q_9, q_7, q_8]$ (from high to low). Six training instances are created in this example.}
\label{tab:E-KPQG}
\end{table*}

\section{KPQG Performance Evaluation}
\label{sec:Experiment}



\subsection{Implementation Details}
We use the \textbf{DeBERTa}$\bf{_{base}}$ \cite{he2020deberta} model for KPQG training. 
The KPQG model is trained by four TITAN V100 GPUs with 10 epochs for 16 hours. In addition, for the QA model for assessing token importance for training data preparation, we use the RACE QA model from \cite{wolf2020huggingfaces}. This model has an accuracy of 84.9\% on the RACE dataset.

\subsection{Human Evaluation}
We use human evaluation to validate the quality of the KPQG model because the premise of the KPQG model allows users to guide the QG direction by indicating keywords expected to be included in the generation result. 300 context paragraphs and the corresponding answers were randomly selected from the test set of EQG-RACE data \cite{jia2020eqg}. We invited 30 evaluators to provide keywords and use the KPQG model to generate questions. The evaluator is asked to compare the difference between QG and KPQG and score [0,1,2] in Likert scale based on the following three metrics:

\begin{itemize}[itemsep= -3 pt,leftmargin= 10 pt, topsep = 0 pt]
\item \textit{\bf Fluency:} how grammar and structural fluency the generated sentence is.
\item \textit{\bf Expectedness:} The extent to which the generated question are in line with expectations.
\item \textit{\bf Answerability:} whether the generated question that can be answered.
\end{itemize}

The human evaluation results are summarized in Table \ref{table:Human evaluation}. We have the following observations. 

For fluency, the two compared models are able to generate grammatical and structural sentences. This is not a surprising result as with the help of language model, the existing QG models are all able to generate fluent question sentences. 

Second, for Expectedness, we see there is a big difference between the two compared models. This result validates the KPQG model addresses the QG controllability concern. 

Thrid, for answerability measure, we also observe improvement. We consider this is due to providing additional keywords guides QG to generate more specific questions other than general questions, which therefore the answerabililty measure is improved.

\begin{table}[ht]
\centering
\resizebox{\linewidth}{!}{
\begin{tabular}{|c|c|c|c|}
\hline
{\bf Model} & {\bf Fluency} &{\bf Expectedness} & {\bf Answerability} \\ 
\hline
{DeBERTa-QG}  & 1.60 & 0.86 & 1.20\\ 
{DeBERTa-KPQG}  & 1.60 & \bf{1.37} &  \bf{1.44}\\   
\hline
\end{tabular}
}
\caption{Human evaluation results}
\label{table:Human evaluation}\vspace{-4mm}
\end{table}

\begin{table*}
\tiny
\centering
\resizebox{\textwidth}{!}{
\begin{tabular}{|p{0.085\textwidth}|p{0.915\columnwidth}|}
\hline 
Example 1 & \\ \hline
Context & At the age of 12, \textbf{Christopher Hirata} already worked on college-level courses, around the time most of us were just in the 7th grade. at the age of 13, this gifted kid became the youngest American to have ever won the gold medal in the International Physics Olympiad. at the age of 16, he was already working with NASA on its project to conquer planet mars. after he was awarded the Ph.D. at Princeton University, he went back to California institute of technology. the next person with very high IQ is Albert Einstein. with an IQ between 160 and 190, Albert Einstein is the genius behind the theory of relativity, which has had great impact on the world of science. he possessed such an amazing ability that after his death, researchers were eager to preserve and make research on his brain in search for clues to his exceptional brilliance, which to this day, has remained a mystery. \\ \hline
Answer & Christopher Hirata  \\ \hline
Gold Question & Who once worked on the project to conquer planet mars? \\ \hline
DeBERTa-QG & 
Who was the youngest American to have ever won the gold medal in the International Physics Olympiad? \\ \hline
Keywords 1 & ``mars'' \\ \hline
DeBERTa-KPQG & Who helped NASA on the project to conquer planet mars? \\ \hline
Keywords 2 & ``mars'', ``who'' \\ \hline
DeBERTa-KPQG & For conquering planet mars, who did he work with NASA? \\ \hline\hline

Example 2 & \\ \hline
Context &  \textbf{Brazil} like the French, Brazilians usually eat a light breakfast. Lunch, the largest meal of the day, usually consists of meat, rice, potatoes, beans and vegetables. between 6:00 p.m. and 8:00 p.m., people enjoy a smaller meal with their families. Brazilians don't mind eating a hurried or light meal and sometimes buy food from street carts. but they always finish eating before walking away. \\ \hline
Answer & Brazil  \\ \hline
Gold Question & In which country do people consider lunch the largest meal? \\ \hline
DeBERTa-QG & 
Which country has a light breakfast? \\ \hline
Keywords 1 & ``largest meal'' \\ \hline
DeBERTa-KPQG & Which country's lunch has the largest meal of the day? \\ \hline
Keywords 2 & ``race'' \\ \hline
DeBERTa-KPQG & Where does lunch usually eat in order of rice, potatoes and vegetables? \\ \hline\hline

Example 3 & \\ \hline
Context &  Three Central Texas men were honored with the Texas department of public safety's director's award in a Tuesday morning ceremony for their heroism in saving the victims of a fiery two car accident. the accident occurred on March 25 when a vehicle lost control while traveling on a rain-soaked state highway 6 near Baylor camp road. it ran into an oncoming vehicle, leaving the occupants trapped inside as both vehicles burst into flames. \textbf{Bonge} was the first on the scene and heard children screaming. he broke through a back window and pulled Mallory Smith, 10, and her sister, Megan Smith, 9, from the wreckage. The girls' mother, Beckie Smith, was not with them at the time of the wreck, as they were traveling with their baby sitter, Lisa Bow Bin. \\ \hline
Answer & Bonge  \\ \hline
Gold Question & Who saved Megan Smith from the damaged car? \\ \hline
DeBERTa-QG & 
Who was the first on the scene and heard children screaming? \\ \hline
Keywords 1 & ``Megan Smith'' \\ \hline
DeBERTa-KPQG & Who saved Megan Smith from the accident? \\ \hline
Keywords 2 & ``which'' \\ \hline
DeBERTa-KPQG & In the accident, which man was the hero of the victims? \\ \hline

\end{tabular}
}
\caption{Case study of KPQG models} 
\label{table:case_study}
\end{table*}

\subsection{Case Study}
We present case study for qualitative evaluation. In Table \ref{table:case_study}, we show three case study. The case studies are selected from the test set of EGQ-RACE \cite{jia2020eqg}. In each case study, we show the context paragraph, answer, and the gold question (the first three row of the tables). We use the gold question to simulate it as the one that the user expects to generate. We list the QG results by DeBERTa-QG and DeBERTa-KPQG with different keyword sets. We have the following discussion for the case studies.

\paragraph{Case 1}
As can be seen from Example 1, although the result of DeBERTa-QG is the correct question, the direction of the question is not the same as the expected golden question. This is because no keywords are used to guide the QG direction. However, in the results of DeBERTa-KPQG, we can see that with the given [``\texttt{mars}''] keyword, the KPQG model has successfully guided the generation toward the golden question. In addition, KPQG can also use keywords to control the generated sentence syntactical structure. For example, in this case, we prompt [``\texttt{mars}'',``\texttt{who}''] for KPQG. We see that ``For conquering plant mars, who did he work with NASA?'' is generated. The generated result not only includes the indicated keywords but also consider the order of the keywords. We consider this ability might be also helpful to improve the QG diversity in terms of different syntactical structure generation.

\paragraph{Case 2}
In Example 2, we can also see that DeBERTa-KPQG's question on the given keyword [``\texttt{largest meat}''] is closer to the golden question. Furthermore, prompting different keywords leads to different results. For example, given the [``\texttt{race}''] keyword, the model generates 
``Where dos lunch usually eat in order of rice, potatoes and vegetables?'', which is a complete different question direction. This result shows that KPQG can control the generation results according to the keywords given by the user. This feature is also helpful for teachers to have inspiration for preparing reading assessment.

\paragraph{Case 3}
Similar to the conclusion from the previous example, in Example 3, we prompt the keyword [``\texttt{Megan Smith}''] to guide the direction of the KPQG model generation. Again, we see the result is close to the golden question. In addition, KPQG can also control the sentence syntax by giving only the ``\texttt{wh-}'' keyword. For example, in Example 3, the answer is that a person’s name usually uses the sentence structure of ``\texttt{who}'', but when the keyword [``\texttt{which}''] is given, KPQG can control the generated result to use ``\texttt{which}'' as a question syntax. This feature can provide users with the specified sentence syntax when generating questions, helping users to have variability and controllability in the application of generating questions.

According to the above results, in addition to proving that our method is simple but effective, it also shows that the controllability and diversity of KPQG are more conducive to practical applications.

\section{Conclusion}
In this paper, we report the following two findings. First, we find that a very simple QG architecture based on pre-trained language models beats the complicated exam-like QG design \cite{jia2020eqg} with or without the keyword indication. Second, by providing keyword information, we can generate results that are closer to the user's expectation. We believe that our method is more practical to the realization of educational QG system applications.

\bibliography{aaai22.bib}

\end{document}